\AtBeginDocument{%
  }
\documentclass[sigconf]{acmart}

\PassOptionsToPackage{table}{xcolor}
\usepackage{xcolor}
\usepackage{booktabs}
\usepackage{multirow}
\usepackage{arydshln}
\usepackage{array} 
\usepackage[utf8]{inputenc}
\usepackage{url}
\usepackage{graphicx}
\usepackage{subcaption}
\usepackage{makecell}
\usepackage{bbding}
\usepackage{pifont}
\usepackage{wasysym}
\usepackage{utfsym}
\usepackage{fontawesome}
\usepackage{tabularx}
\usepackage{enumitem}

\acmConference[MM '25]{ACM International Conference on Multimedia}{October 27--31, 2025}{Dublin, Ireland}
\renewcommand\footnotetextcopyrightpermission[1]{}
\usepackage{enumitem}
\definecolor{skyblue}{RGB}{	70 ,130 ,180} % 这是常用的天蓝色 RGB 值
\definecolor{lightblue}{RGB}{135 ,206 ,250}
\begin{document}

\title{RecipeGen: A Step-Aligned Multimodal Benchmark for Real-World Recipe Generation}
% \author{Cheng Zhang}
% \email{zhangcheng2122@mails.jlu.edu.cn}
% \affiliation{%
%   \institution{Jilin University}
%   \city{Changchun}
%   \country{China}
% }
\author{Ruoxuan Zhang}
\email{zhangrx22@mails.jlu.edu.cn}
\affiliation{%
  \institution{Jilin University}
  \city{Changchun}
  \country{China}
}

\author{Jidong Gao}
\email{3122008925@mail2.gdut.edu.cn}
\affiliation{%
  \institution{Guangdong University of Technology}
  \city{Guangzhou}
  \country{China}
}

\author{Bin Wen}
\email{wenbin2122@mails.jlu.edu.cn}
\affiliation{%
  \institution{Jilin University}
  \city{Changchun}
  \country{China}
}

\author{Hongxia Xie}
\email{hongxiaxie@jlu.edu.cn}
\affiliation{%
  \institution{Jilin University}
  \city{Changchun}
  \country{China}
}

\author{Chenming Zhang}
\email{zhangcm9921@mails.jlu.edu.cn}
\affiliation{%
  \institution{Jilin University}
  \city{Changchun}
  \country{China}
}

\author{Hong-Han Shuai}
\email{hhshuai@nycu.edu.tw}
\affiliation{%
  \institution{National Chiao Tung University}
  \city{Taipei}
  \country{Taiwan}
}

\author{Wen-Huang Cheng}
\email{wenhuang@csie.ntu.edu.tw}
\affiliation{%
  \institution{National Taiwan University}
  \city{Taipei}
  \country{Taiwan}
}
\begin{abstract}
Creating recipe images is a key challenge in food computing, with applications in culinary education and multimodal recipe assistants. However, existing datasets lack fine-grained alignment between recipe goals, step-wise instructions, and visual content. We present RecipeGen, the first large-scale, real-world benchmark for recipe-based Text-to-Image (T2I), Image-to-Video (I2V), and Text-to-Video (T2V) generation. RecipeGen contains 26,453 recipes, 196,724 images, and 4,491 videos, covering diverse ingredients, cooking procedures, styles, and dish types. We further propose domain-specific evaluation metrics to assess ingredient fidelity and interaction modeling, benchmark representative T2I, I2V, and T2V models, and provide insights for future recipe generation models. Project page is available at  \href{https://wenbin08.github.io/RecipeGen}{\textcolor{skyblue}{https://wenbin08.github.io/RecipeGen}}.

\end{abstract}

%    <concept>
%        <concept_id>10010147.10010178.10010224.10010225</concept_id>
%        <concept_desc>Computing methodologies~Computer vision tasks</concept_desc>
%        <concept_significance>500</concept_significance>
%        </concept>
%  </ccs2012>

% \ccsdesc[500]{Computing methodologies~Computer vision tasks}

%    <concept>
%        <concept_id>10010405</concept_id>
%        <concept_desc>Applied computing</concept_desc>
%        <concept_significance>300</concept_significance>
%        </concept>
%    <concept>
%        <concept_id>10010405.10010476</concept_id>
%        <concept_desc>Applied computing~Computers in other domains</concept_desc>
%        <concept_significance>500</concept_significance>
%        </concept>
%  </ccs2012>

% \ccsdesc[300]{Applied computing}
% \ccsdesc[500]{Applied computing~Computers in other domains}

\begin{teaserfigure}
  \includegraphics[width=\textwidth]{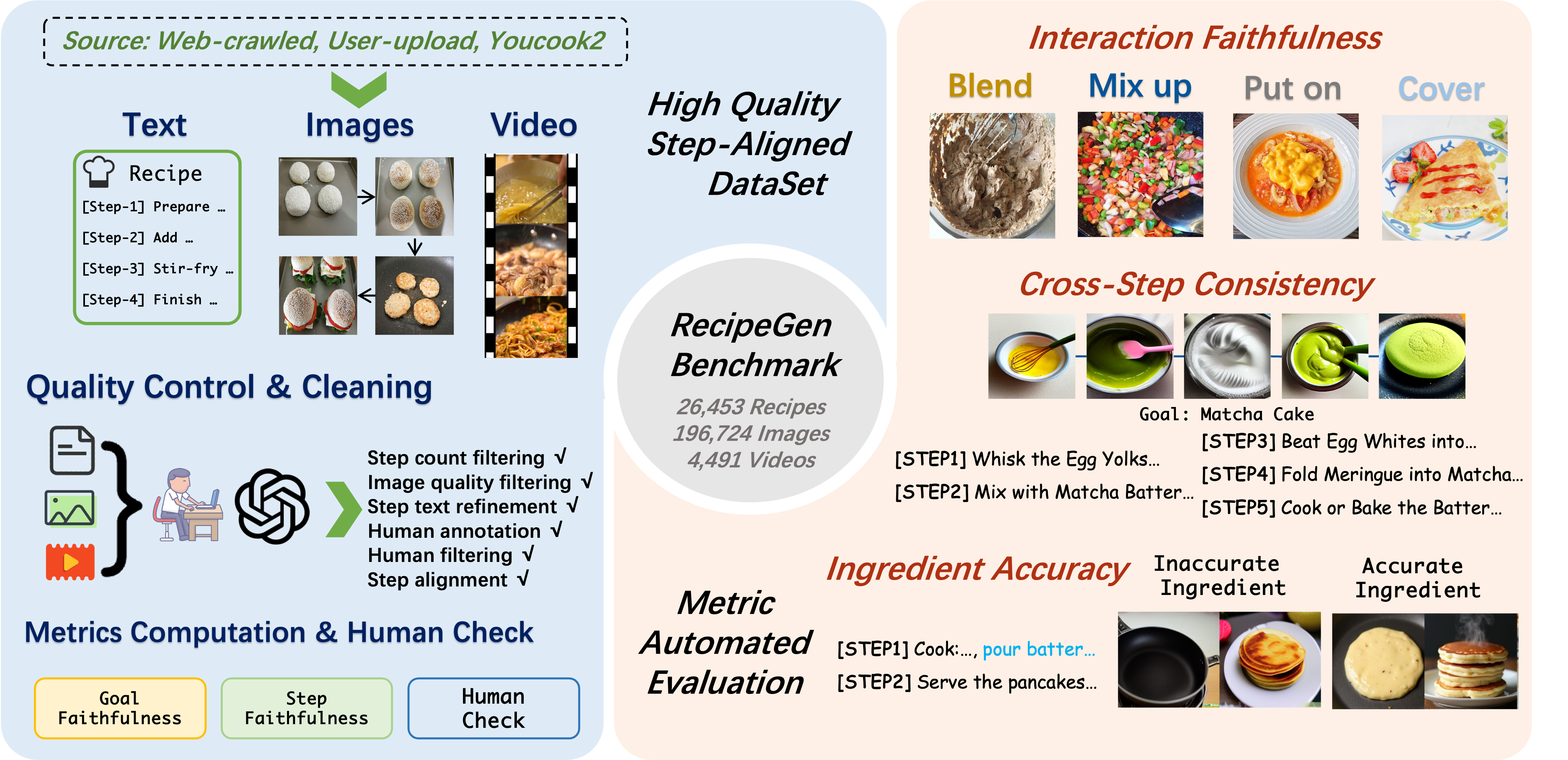}
  \caption{RecipeGen Benchmark Overview.  A comprehensive real-world benchmark for multimodal recipe generation, supporting consistent and realistic image/video synthesis. It integrates an automatic and manual quality control pipeline and introduces domain-specific evaluation metrics to assess generation fidelity and usability.}
  % \Description{Enjoying the baseball game from the third-base
  % seats. Ichiro Suzuki preparing to bat.}
  \label{fig:fig1}
\end{teaserfigure}


\maketitle

\section{Introduction}
\begin{center}
    \vspace*{2em}
    \emph{``Food is the paramount necessity of the people.''}
\end{center}

Generating illustrated or video-based instructions has attracted increasing attention as an effective way to guide users through complex, step-by-step tasks. Whether assembling furniture, repairing devices, or executing cooking procedures, visual representations—especially temporal ones—can significantly reduce user confusion and improve task success rates.
One particularly compelling application is \textbf{Recipe Generation}, where step-by-step illustrations enhance the interpretability and accessibility of cooking instructions. Existing studies in recipe image generation can be categorized into creating dish images from the recipe title and ingredients~\cite{h2020recipegpt}, from recipe steps and ingredients~\cite{pan2020chefgan, liu2023ml, han2020cookgan}, or conditioned on a reference dish image~\cite{chhikara2024fire, dosovitskiy2020image, wang2020structure, wang2022learning}. While these approaches have shown promise, they often struggle to generate intermediate steps with clarity and consistency, which leads to confusion and misinterpretation. A primary reason for this limitation lies in existing datasets~\cite{marin2021recipe1m+, bien-etal-2020-recipenlg, batra2020recipedb}, which typically only include final images of the dish, lacking detailed snapshots that capture essential transitions between steps.

To advance research in recipe generation, we present a novel multimodal dataset named \textbf{RecipeGen} specifically designed for both recipe image and video generation tasks. Our dataset spans a broad spectrum of cuisines, cooking techniques, and dish categories, with meticulous step-level annotations comprising detailed textual descriptions paired with corresponding visual content. It consists of two well-aligned subsets: (1) \textbf{a large-scale text-image collection} featuring per-step visual illustrations, and (2) \textbf{a curated text-image-video subset} that includes step-aligned cooking videos alongside manually selected keyframes for each step. This dual-modality design enables unified training paradigms and facilitates cross-modal evaluation of generative models under both static and dynamic visual supervision.

Cooking procedures inherently involve complex ingredient transformations and intricate inter-ingredient interactions, posing significant challenges to maintaining semantic and visual consistency across steps. Furthermore, existing evaluation metrics largely fail to capture these fine-grained dynamics and cross-step coherency. To fill this gap, we propose a comprehensive evaluation framework consisting of three novel metrics: \textbf{Cross-Step Consistency}, which quantifies the visual coherence between consecutive steps; \textbf{Ingredient Accuracy}, which measures the fidelity of generated ingredients relative to ground-truth annotations; and \textbf{Interaction Faithfulness,} which assesses the correctness of depicted ingredient interactions in accordance with the described cooking actions. Together, these metrics offer a more task-specific and holistic assessment of recipe generation quality.

\noindent \textbf{Our contributions are threefold:}
\begin{itemize}[leftmargin=1em] 
\item \textbf{High-quality step-aligned multimodal data.} RecipeGen includes both text-image and text-image-video subsets, supporting a variety of generation tasks such as text-to-image, image-to-video, and text-to-video. It offers detailed, step-level textual and visual annotations with minimal non-instructional content, enabling fine-grained semantic and visual understanding of complex cooking procedures.
\item \textbf{Large-scale and diverse dataset coverage.} RecipeGen encompasses a wide array of ingredients, cooking techniques, and regional cuisines, supporting the development of robust, generalizable recipe understanding and generation models.

\item \textbf{A novel and comprehensive evaluation suite.} We introduce a hierarchical evaluation protocol with three new metrics—Cross-Step Consistency, Ingredient Accuracy, and Interaction Faithfulness—that better capture procedural correctness and semantic-visual alignment in generated recipes.

\end{itemize}

\section{Related Work}

Food is closely related to people's lives, and with the development of human society, diets have diversified. To facilitate the management of human life and health, the field of food computing  \cite{min2019survey} has emerged. Academically, topics such as  food segmentation \cite{lan2023foodsam,yin2023foodlmm}, food recognition \cite{min2023large,8779586,zhang2023deep}, food recommendation \cite{min2019food,wang2021market2dish}, food reasoning \cite{zhou2024foodsky} and recipe image generation have gained significant attention.

\textbf{Food Computing Dataset.}
There are several existing datasets for food computing tasks. For instance, Food-101 \cite{bossard2014food}, VireoFood-172 \cite{chen2016deep}, Recipes242k \cite{rokicki2018impact}. However, these datasets are primarily used for fine-grained tasks such as food or ingredients classification, nutritional analysis, calorie calculation, etc. In addition to these information, FoodEarth \cite{zhou2024foodsky} utilizes LLMs to develop a comprehensive dataset for food and nutrition analysis. Similarly, Recipe1M \cite{marin2021recipe1m+} provides recipe instructions linked to each final dish image. However, neither dataset includes images for each individual recipe step. While the recipe section of VGSI \cite{zhou2018towards} offers step images for each instruction, many of these images are presented in a comic style. To address this gap, we propose the RecipeGen Benchmark, which is the first dataset that provides a real-world step-image recipe dataset. 
% We will provide detailed statistic comparison in section \ref{sec:benchmark}. 

\section{RecipeGen Benchmark}
\label{sec:benchmark}

\begin{figure*}[htbp]
  \centering
  \begin{subfigure}[b]{0.45\textwidth}
    \includegraphics[width=\textwidth]{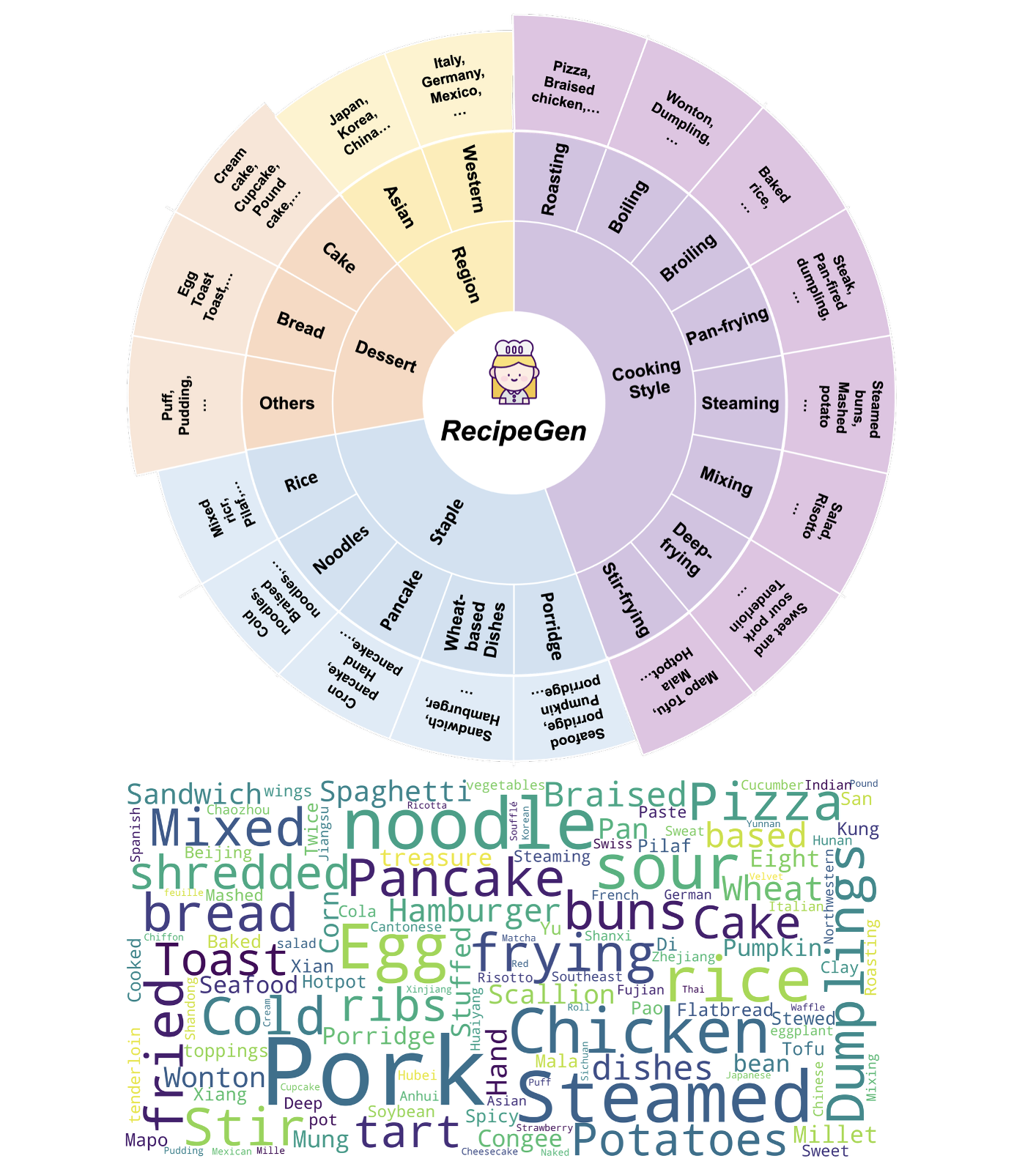}
    \caption{\textbf{Top}: The distribution of some keywords in RecipeGen Benchmark. \textbf{Bottom}: Word cloud of recipe text queries shows a considerable degree of diversity}
    \label{fig:sub1}
  \end{subfigure}
  \hfill
  \begin{subfigure}[b]{0.5\textwidth}
    \includegraphics[width=\textwidth]{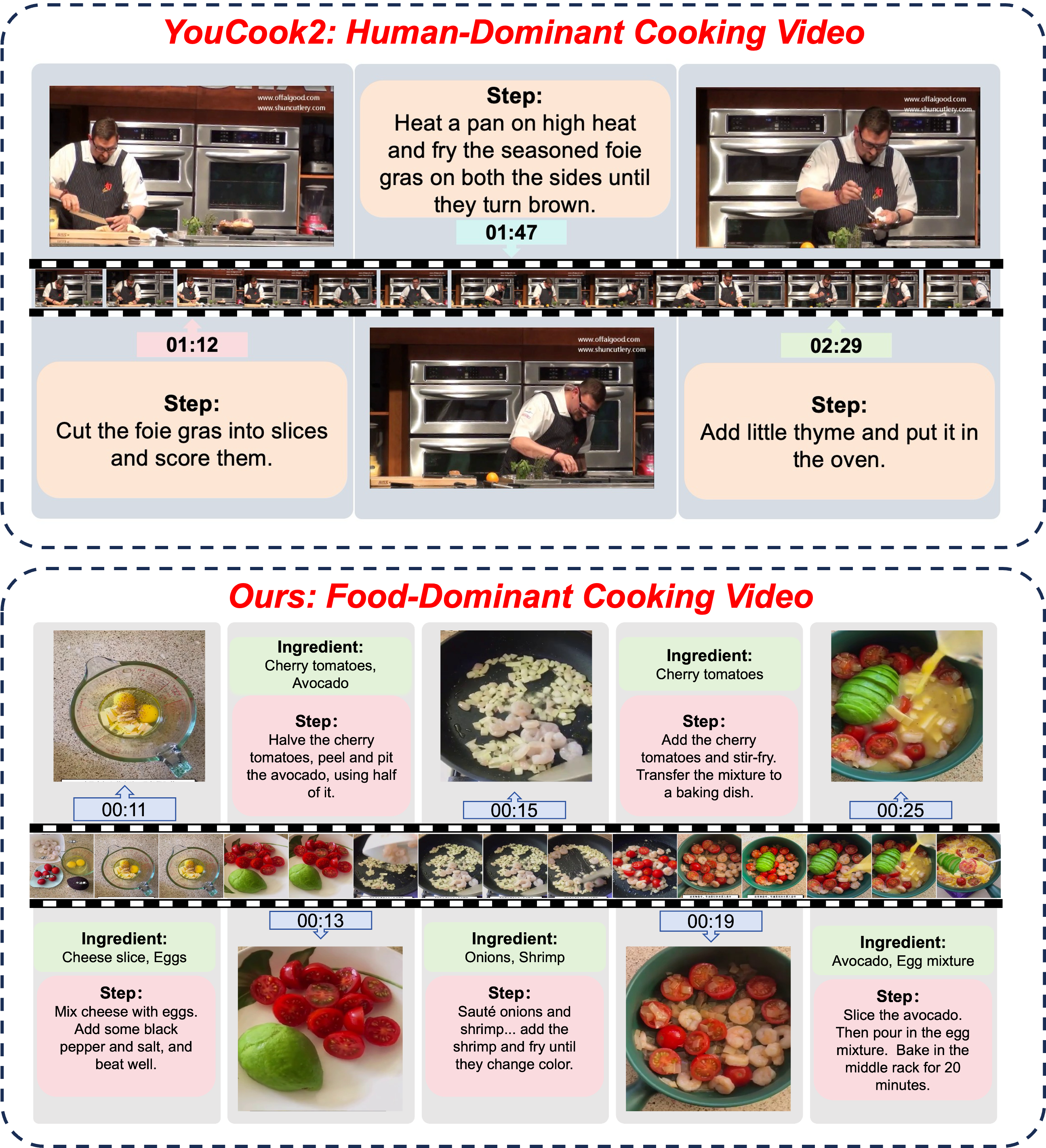}
    \caption{Comparison with YouCook2.  RecipeGen focuses on food-centric videos, making it more suitable for visualizing cooking procedures.}
    \label{fig:sub2}
  \end{subfigure}
  \caption{Overview of Our RecipeGen Benchmark.}
  \label{fig:main}
\end{figure*}

% \begin{figure*}[t]
%   \centering
 
%   \includegraphics[width=1\linewidth]{fig/data.png} 

%   \caption{Some examples in RecipeGen. Each sample contains its goal, steps, and the images corresponding to each step.
% }
%   \label{fig:example}
% \end{figure*}

% \begin{table*}[h!]
%   \centering
%   % \tabcolsep=4.5pt
%   % \fontsize{7}{11}\selectfont
%   \begin{tabular}{l c c c c c c c c } 
%     \hline
%     \multirow{2}{*}{\textbf{DataSet}} & \multirow{2}{*}{\textbf{Recipes}} & \multirow{2}{*}{\textbf{Steps/Images}}  &\multirow{2}{*}{\textbf{Video}}& \multirow{2}{*}{\textbf{Modality}} &  \multirow{2}{*}{\textbf{Task}}&\multirow{2}{*}{\textbf{Source}} &\multicolumn{2}{c}{\textbf{Data Faithfulness}}  \\ 
% \cline{8-9}
%           &          &         &&          &        && \textbf{GF} & \textbf{SF} \\ 
%     \hline
%     VGSI-Recipe &1,157&6,417 && T, I&  T2I&wikiHow & 27.35& 29.03\\ 
%     YouCook2 & 2,000 & 15,433  && T, V&  VLU&YouTube & 27.77& 29.93\\ 
%     \textbf{RecipeGen (Ours)} & \textbf{21,944} & \textbf{139,872}  && \textbf{T, I, V}&  \textbf{T2I, I2V, T2V}&\textbf{douguo,YouTube}
%     & & \\ 
%     \hline
%   \end{tabular}
%   \caption{Comparison among Different Recipe Datasets. }
%   \label{tab:RecipeDataset}
% \end{table*}

\begin{table*}[t]
  \centering
  \small
  \setlength{\tabcolsep}{5pt}
  \renewcommand{\arraystretch}{1.3}
  \begin{tabular}{lccccccccc}
    \toprule
    \multirow{2}{*}{\textbf{Dataset}} & 
    \multirow{2}{*}{\textbf{\#Recipes}} & 
    \multirow{2}{*}{\textbf{Images}} & 
    \multirow{2}{*}{\textbf{Videos}} & 
    \multirow{2}{*}{\textbf{Modalities}} &  
    \multirow{2}{*}{\textbf{Tasks}} & 
    \multirow{2}{*}{\textbf{Source}} & 
    \multicolumn{2}{c}{\textbf{Data Faithfulness}} \\ 
    \cmidrule(l){8-9}
    & & & & & & & \textbf{Goal (GF)} & \textbf{Step (SF)} \\ 
    \midrule
    \texttt{VGSI-Recipe}~\cite{yang2021visual} & 1,157 & 6,417 & 0 & T, I & T2I & wikiHow & 27.35 & 29.03 \\ 
    \texttt{YouCook2}~\cite{zhou2017procnets} & 2,000 & 0 & 2,000 & T, V & VU & YouTube & 27.77 & 29.93 \\ 
    \texttt{\textbf{RecipeGen}} & \textbf{26,453} & \textbf{196,724} & \textbf{4,491} & \textbf{T, I, V} & \textbf{T2I, I2V, T2V} & \textbf{Douguo, YouTube, YouCook2} & \textbf{28.94} & \textbf{30.31} \\
    \bottomrule
  \end{tabular}
  \caption{Comparison of \texttt{RecipeGen} with existing recipe-related datasets in terms of scale, modalities, supported tasks, and data faithfulness. }
  \label{tab:RecipeDataset}
\end{table*}

The diversity of human culinary practices, developed over centuries, has led to a vast array of ingredient combinations and cooking methods, posing challenges for constructing comprehensive recipe step-image datasets. There are few existing datasets that cover a wide range of ingredients and interactions, and step counts are often limited.

To address the limitations of existing recipe datasets—such as limited step coverage, narrow ingredient diversity, and a predominant focus on Western cuisine—we introduce the \textbf{RecipeGen Benchmark}, a large-scale multimodal dataset containing both text-image and text-image-video recipes. RecipeGen comprises 26,453 curated recipes, paired with 196,724 step-aligned images and 4,491 cooking videos, covering a wide spectrum of regional dishes and culinary styles. The dataset is constructed from user-uploaded platforms, web-crawled resources, and existing datasets, offering a robust foundation for evaluating Text-to-Image (T2I), Image-to-Video (I2V), and Text-to-Video (T2V) generation models.
Unlike previous datasets such as VGSI-Recipe \cite{yang2021visual} and YouCook2 \cite{zhou2017procnets}, which primarily focus on Western cuisine, RecipeGen captures a broader spectrum of culinary traditions. For instance, the same ingredient—beef—may be pan-fried as a steak in Western recipes, but slow-braised with root vegetables in East Asian dishes. By incorporating these regional variations, RecipeGen facilitates a greater understanding of diverse ingredient treatments and cross-cultural cooking strategies.

% To address this gap, we propose \textbf{RecipeGen Benchmark (RecipeGen)}, a goal-instruction image dataset designed to evaluate text-to-image models (T2Is) in understanding cooking instructions. RecipeGen includes 21,944 recipes with 139,872 images and steps, encompassing diverse regional dishes and cooking styles. Collected from a large set of user-uploaded, real-world recipes\footnote{\url{www.douguo.com}}, RecipeGen provides a solid foundation for the evaluation of the T2I model. This dataset captures the distinctive ways different culinary traditions prepare similar ingredients and combines unique cooking techniques. For instance, Western recipes may pan-fry beef as a steak, while East Asian cuisine might braise it with root vegetables. Unlike existing datasets focused primarily on Western cuisines, such as VGSI-Recipe \cite{yang2021visual}, RecipeGen spans a wide range of regional styles, including East and Southeast Asian cuisines, enhancing model understanding of varied preparation techniques. 

\subsection{Construction Procedure for Image-Text Recipes}

To ensure RecipeGen’s diversity, we curate 158 keywords across different cuisines (e.g., Cantonese, Mexican), dish types (e.g., main courses, desserts), and cooking techniques (e.g., stir-frying, roasting), gathering 21,944 recipes.   
% \footnote{After Quality Control, the number of recipe is 21,944 and the number of images is 139,872.}
% \noindent \textbf{Steps.} RecipeGen reflects a variety of cooking methods, capturing simple recipes with minimal steps and complex dishes with multiple preparation stages, accommodating diverse user techniques.
% \cref{fig:keywords_distribution} shows distribution of keayworks. 
% \noindent \textbf{Ingredients.} RecipeGen features an extensive range of ingredients contributed by users from different regions, allowing for a comprehensive representation of various flavor profiles and ingredient interactions.
% \textbf{Keyword Selection and Coverage.}
% \label{keywords}

\textbf{Keyword Selection and Coverage.}
A total of 158 keywords covering 158 different cuisines, cooking methods, and dish types are defined to ensure broad and structured coverage. Fig.~\ref{fig:main} illustrates their distribution. Based on these keywords, we collect a total of 29,026 recipes. 

 \textbf{Quality Control.}
To maintain consistency and clarity, we implement a quality control process to address common issues like incomplete instructions or overly detailed steps. This process involved: (1) Removing recipes with low quality (including lack images or steps, mismatched numbers of images and steps, vague directions \footnote{For example, some steps in the recipe only contain phrases like ``see as images" without providing clear or actionable instructions.}). In this step, we delete 4,978 recipes; (2) Using GPT-4o \cite{achiam2023gpt} to clean and refine recipes by eliminating irrelevant content \footnote{For instance, some users include exclamatory phrases such as ``This tastes amazing!" or suggestions like ``Using ingredient A instead of B works better."}, merging redundant steps, and summarizing for coherence. And then, we remove recipes where the LLM's hallucination issues caused a single step to be split into two, as well as recipes with incorrect output formats. A total of 2,104 recipes were deleted. After Quality Control, the number of recipes is 21,944 and the number of images is 139,872; (3) Computing data faithfulness and using human check to insure the quality. The result of metrics is in Tab. \ref{tab:RecipeDataset}. 

% \begin{figure*}[t]
%   \centering
 
%   \includegraphics[width=1\linewidth]{fig/quality_control.png} 

%   \caption{Quality Control Prompt for GPT-4o. We utilize GPT-4o to merge steps and generate captions.}
%   \label{fig:quality_control}
% \end{figure*}

% \textbf{Prompts for GPT-4o.} To ensure Quality Control, we employ GPT-4o to combine adjacent simple steps, generate captions to prevent over-merging by GPT-4o and translate the output into English. We establish clear principles for GPT-4o and outline a step-by-step process to guide it in producing accurate and appropriate results. 

\textbf{Metrics Computation and Human Check.} To assess dataset quality, we use three metrics. \textbf{Goal Faithfulness (GF)} measures the CLIP score between the final image and its caption, reflecting dish completeness. \textbf{Step Faithfulness (SF)} computes the CLIP score between each image and its corresponding step. As shown in Tab.~\ref{tab:RecipeDataset}, our RecipeGen dataset outperforms VGSI-Recipe and YouCook2 across all metrics, with GF of 28.94, SF of 30.31.
 In addition to evaluating the dataset quality using step faithfulness and goal faithfulness metrics, we also conduct a human review. Specifically, annotators are asked to verify three aspects: (1) whether the recipe matches its dish name, (2) whether each step corresponds correctly to its associated image, and (3) whether the images within the same recipe exhibit visual continuity. We randomly sample 2,000 recipes, covering more than 12,000 images for this review. Among these samples, we identify 12 instances where the dish name was mislabeled (e.g., ``labeling dumplings'' as ``steamed buns''), and 7 instances where all the textual steps were generic and lacked specific ingredient descriptions (e.g., using phrases like ``as shown in the image'' or ``ready to serve''). As well as 6 recipes in which no pictures existed and 17 cases that lacked pictures corresponding to some of the steps.
% RecipeGen’s broader range of steps and greater diversity in regional cuisines make it a robust resource for T2I model evaluation and training, supporting more accurate and diverse recipe instruction generation.

 %In terms of cross-image consistency (CIC), although our dataset contains more average steps per recipe than VGSI-Recipe, we maintain a higher consistency across consecutive step images within the same recipe. 

% \begin{figure}[t]
%   \centering
 
%   \includegraphics[width=0.9\linewidth]{fig/data_main_food.png} 

%   \caption{Various staples keywords 
% in our proposed RecipeGen Benchmark.  Our staple food keywords are diverse and can be categorized into five main groups.}
%   \label{fig:staple_distribution}
% \end{figure}

% \begin{figure}[t]
%   \centering
 
%   \includegraphics[width=1\linewidth]{fig/gpt4o_for_data.png} 

%   \caption{Dataset Construction Procedure. We first analyze the characteristics of the dishes and select 158 keywords. Subsequently, we utilize GPT-4o to perform quality control by omitting irrelevant steps, merging adjacent simple actions, and generating captions. Finally, we calculate metrics and conduct human checks to ensure the usability of the dataset.
% }
%   \label{fig:gpt-benc}
% \end{figure}

\subsection{Construction Procedure for Image-Text-Video Recipes}

For the video data, the sources primarily consist of web-crawled videos and a curated subset of the YouCook2 \cite{zhou2017procnets} dataset. In total, RecipeGen includes 4,491 recipes paired with instructional videos.

\textbf{Problem of Existing Dataset.}
YouCook2 is a dataset collected from YouTube that contains 2,000 cooking videos with temporal annotations for each step \footnote{We only utilize the training and validation sets, as these are the only subsets that provide detailed step-wise annotations necessary for our processing. In total, these two splits contain 1,774 recipe videos.}, including start and end timestamps, along with corresponding video segments. However, the dataset presents several limitations for step-level recipe generation and food-centric modeling. First, as shown in Fig. 1, \textbf{because YouCook2 was designed for general video captioning rather than recipe-specific tasks, and its videos are often sourced from lifestyle vloggers, the resulting content is typically human-dominant rather than food-dominant, with limited focus on ingredients and cooking actions}. To quantify this issue, we use YOLOv8~\cite{Jocher_Ultralytics_YOLO_2023} to estimate the temporal proportion of frames containing people. Sampling one frame every 10 frames and applying a confidence threshold of 0.5 for the “person” label, we find that the average proportion of person-visible duration across the dataset is 46.58\%, with 850 videos having more than 50\% of their duration focused on people. This significantly reduces the visual relevance for food-centric tasks. Second, because each step is represented as a video segment rather than an individual image, \textbf{YouCook2 lacks explicit step-level image modality}, limiting its applicability for tasks such as step-wise image generation or vision-language alignment at the step level.

\textbf{Quality Control for YouCook2.}
To adapt YouCook2 for recipe video generation, we applied several preprocessing steps. First, we used YOLOv8 (confidence > 0.5) to detect humans in sampled frames (1 every 10 frames). Videos with over 50\% person presence were discarded; for the rest, person-containing frames were removed and clips reassembled. This yielded 924 videos, reducing person-visible content to just 0.98\%, ensuring focus on cooking actions and ingredients.

% To adapt the YouCook2 dataset for our recipe video generation task, we applied a series of preprocessing and filtering steps to address its limitations. First, to detect and quantify the presence of people in the videos, we employ YOLOv8 with the “person” class and a confidence threshold of 0.5, sampling one keyframe every 10 frames. Videos in which over 50\% of the sampled frames contained humans were discarded. For the remaining videos, all frames containing people were removed and the remaining clips were stitched back together. This process result in a refined set of 924 videos. After processing, the average proportion of person-visible duration across the dataset is reduced to just 0.98\%, ensuring that the visual content is focused on cooking actions and ingredients rather than hosts or background subjects.
Second, since each annotated step in YouCook2 corresponds to a continuous video segment rather than a single representative image, we extract a keyframe for each step to enable image-based modeling. Specifically, for each step, we compute the CLIP similarity between the step’s textual description and all frames within the corresponding video segment. The frame with the highest CLIP similarity score is selected as the keyframe representing that step. Finally, a total of 924 recipes are retained.

\textbf{Quality Control for Web-Crawled Data.}
We expanded the dataset by collecting recipe videos from Douguo\footnote{\url{www.douguo.com}} and YouTube\footnote{\url{www.youtube.com}}. Douguo offers aligned step-by-step instructions, images, and videos, requiring no additional processing. YouTube lacks explicit step annotations but provides auto-generated “key moments,” which we extract via lazy loading during crawling.
As the associated thumbnails are often low-resolution, we enhance quality by sampling frames around each timestamp and selecting the one with the highest SSIM~\cite{wang2004image} score to replace the thumbnail, ensuring visual clarity and semantic consistency.

% We further expand our dataset by collecting additional recipe videos from Douguo\footnote{\url{www.douguo.com}} and YouTube\footnote{\url{www.youtube.com}}. Douguo provides user-uploaded content that includes step-by-step instructions, videos, and images, all aligned by the creators. This eliminates the need for us to manually extract and align individual steps. In contrast, YouTube videos typically lack explicit step annotations and associated images. However, YouTube offers automatically generated “key moments” that mark important segments of a video. We leverage this functionality by triggering lazy loading to extract these key moments during crawling.
% The thumbnail generated from these key moments are often low in resolution. To ensure image quality, we implement a refinement process: for each thumbnail, we sample one frame every five frames around its timestamp from the original video and compute the Structural Similarity Index (SSIM) \cite{wang2004image} between each sampled frame and the low-resolution thumbnail. The frame with the highest SSIM score is selected as the high-resolution replacement. This ensures that all collected frames are both visually clear and semantically consistent with the original key moment.

To further improve data quality, we apply a two-stage filtering process:

1) \textbf{Automated Text Cleaning}: A GPT-based method removes non-informative phrases and irrelevant labels common in user-generated content.

2) \textbf{Manual Alignment}: We manually correct mismatches between cleaned text and images to maintain accurate step-image pairs.  After this process, a total of 3,567 recipes are retained.

\subsection{Metrics}

Cooking is an inherently complex, multi-step process that involves \textbf{sequential actions}, \textbf{ingredient interactions}, and \textbf{temporal dependencies}. To faithfully evaluate model performance in such a setting, we design metrics along two primary dimensions, , as summarized in Tab.\ref{tab:evaluation-metrics}:

\textbf{Step-Level Faithfulness.} At the step level, each generated image is evaluated for its alignment with the corresponding textual instruction in two aspects: \textbf{ingredient accuracy}, which measures whether all required ingredients are correctly and completely depicted, and \textbf{interaction accuracy}, which assesses whether the visualized ingredient interactions match the semantics described in the step. We categorize interaction types into five visual patterns: (1) \textbf{Mix up}—multiple ingredients appear together while retaining their original form (e.g., tomato and scrambled egg); (2) \textbf{Blend}—ingredients are fully fused into a new form with no distinct identity (e.g., red bean paste mixed into dough); (3) \textbf{Put on}—one ingredient is simply placed atop another without merging (e.g., cheese on pizza); (4) \textbf{Cover}—one ingredient fully wraps or encloses another (e.g., dumpling wrapper enclosing filling); and (5) \textbf{No relationship}—no interaction occurs between ingredients, typically during early-stage preparations such as cutting or washing. Detailed example is in Fig.\ref{fig:fig1}.
%For each generated image, we finetune a multimodal model (e.g., Qwen2.5-VL) to recognize both ingredients and interaction types, and compare them with the reference annotations for accuracy evaluation.

\textbf{Cross-Step Consistency.} Beyond individual steps, we assess Cross-Step Consistency by measuring whether visual attributes of the same ingredients are maintained across steps and whether the total number of generated steps aligns with the ground truth, using visual $l_2$ distance and step count difference as indicators. 
%These metrics collectively capture the fine-grained visual-semantic alignment and procedural coherence crucial for recipe-oriented image generation tasks.

\begin{table}[h]
\centering
\small % 或 \scriptsize 如果还需要更小字体
\renewcommand{\arraystretch}{1.5}
\caption{Evaluation Metrics for Recipe Generation.}
\begin{tabularx}{\linewidth}{p{2cm} X}
\toprule
\textbf{Metric} & \textbf{Description} \\
\midrule

\textbf{Goal \newline Faithfulness} & CLIP~\cite{radford2021learning} similarity between the final image and last-step caption, measuring alignment with the overall goal. \\
\textbf{Step \newline Faithfulness} & Assesses each image’s alignment with its step caption using CLIP.\\
\textbf{Ingredient\newline  Accuracy } & Uses GPT-4o \cite{achiam2023gpt} to extracts step-wise ingredients and finetune Qwen2.5-VL \cite{Qwen2.5-VL} for three epochs to predict them from generated images. The trained model serves as a VQA model to detect the presence of expected ingredients, with the ratio of correctly detected to ground-truth ingredients as accuracy.
\\
\textbf{Interaction \newline Faithfulness
} & Based on different interaction types—Mix up, Blend, Put on, Cover, and No relationship—we design a standardized prompt that guides GPT-4o to generate one-sentence captions grounded in visible features such as color, texture, spatial layout, and perceived fusion. We then compute the CLIP similarity score between each generated image and its corresponding caption to evaluate interaction consistency.
\\
\textbf{Cross-Step \newline Consistency} & Based on StackedDiffusion~\cite{menon2024generating} and DINOv2~\cite{dino}, uses $l_2$ distance and step count difference to assess visual and numerical consistency. \\

\bottomrule
\end{tabularx}

\label{tab:evaluation-metrics}
\end{table}

\subsection{Dataset Statistics}

Overall, our proposed RecipeGen Benchmark has four notable features:
\begin{itemize}[leftmargin=1em]  % 调整这里的 1em 可控制缩进宽度
    \item \textbf{Broad Step Distribution and Ingredient Diversity}: The dataset includes recipes with 2 to 25 steps, with an average of 7.4 steps, supporting the modeling of long-range semantic relationships. RecipeGen provides an average of 9 ingredients per recipe (including seasonings), capturing a rich variety of ingredients and interactions.

    \item \textbf{Variety of Cooking Styles}: The dataset encompasses a wide range of cooking styles enhanced by numerous unique keywords, making it versatile across different culinary processes.

  \item \textbf{High-quality step-aligned multimodal content}: Our dataset contains cooking videos with a high proportion of effective instructional frames, significantly reducing non-informative host appearances. Each video is accompanied by detailed, step-by-step textual instructions and corresponding diagrammatic illustrations, which together enable a fine-grained understanding of the cooking process.

\item \textbf{Comprehensive evaluation metrics}: We design evaluation metrics that capture both the overall visual quality and fine-grained semantic accuracy. These metrics assess not only the coherence and realism of the generated content at a high level, but also verify the presence, correctness, and interactions of individual ingredients throughout each recipe step.

    \item \textbf{Real-World Representativeness}: Collected from various users, RecipeGen consists entirely of real-world recipes and closely resembles the types of instructions that users might upload in practical scenarios. Quality control using GPT-4o ensures that steps are trustworthy, reflecting actual cooking practices.
\end{itemize}

\section{Experiment}
We primarily evaluate the performance of image generation and video generation. We also evaluate various multimodal large models on fine-grained classification and recipe question answering tasks; details can be found in \href{https://wenbin08.github.io/RecipeGen/}{\textcolor{skyblue}{Project page}}.

\begin{figure*}[t]
  \centering
 
  \includegraphics[width=1\linewidth]{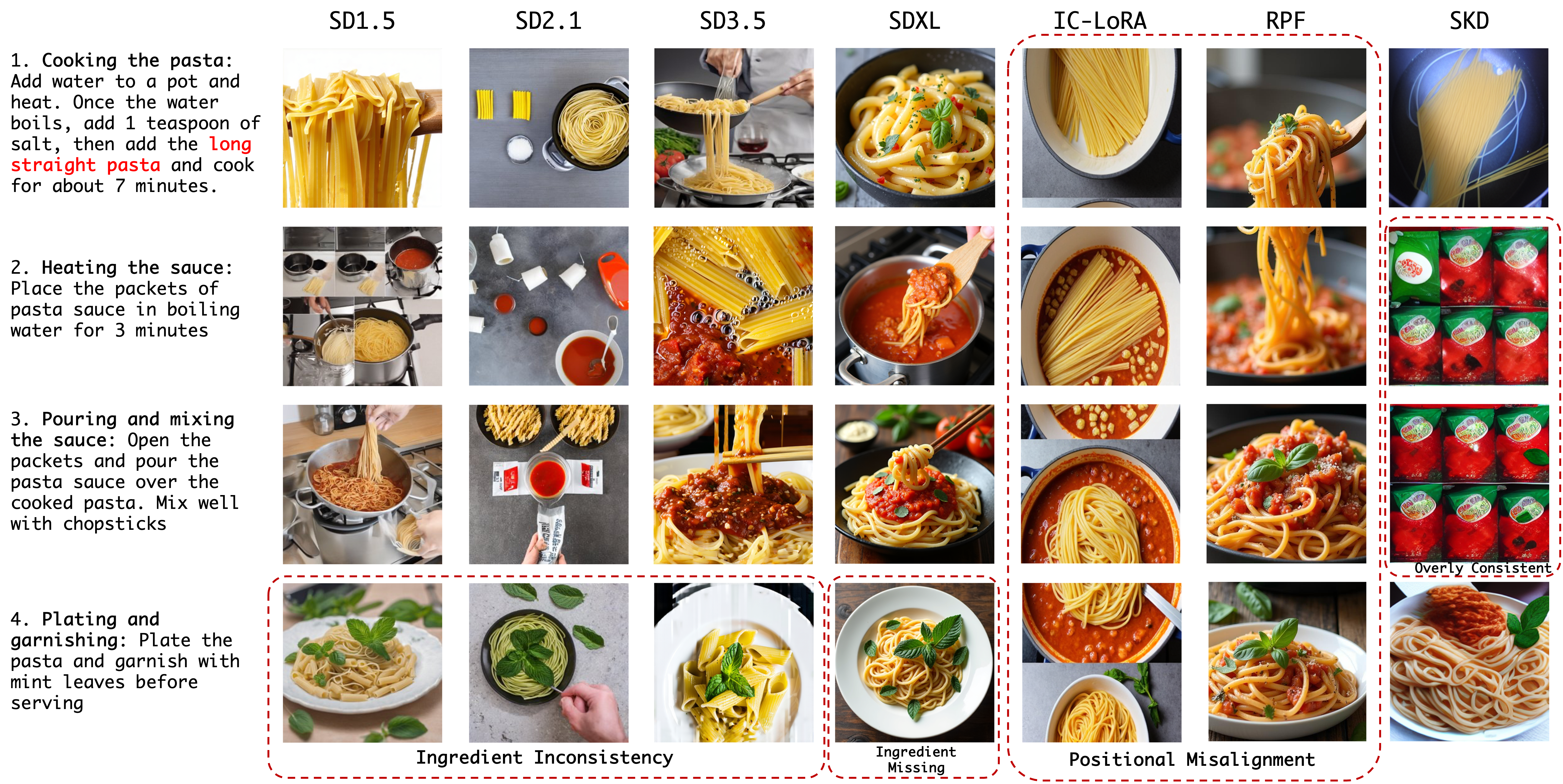} 

  \caption{Visualization of T2I models on Recipe Image Generation task.
}
  \label{fig:vis}
\end{figure*}

\subsection{Text to Image Generation}
In this section, we generate an image for each step in a recipe by feeding the step-wise textual descriptions into various text-to-image generation models. We explore both UNet-based and DiT-based architectures to assess their performance in generating coherent and semantically faithful images for procedural cooking steps. Specifically, the UNet-based models include \textit{Stable Diffusion v1.5 (SD1.5)}  \cite{rombach2022high}, \textit{Stable Diffusion v2.1 (SD2.1)} \cite{stabilityai2022sd21}, \textit{Stable Diffusion XL (SDXL)} \cite{podell2023sdxl}, and \textit{StackedDiffusion (SKD)} \cite{menon2024generating} , while the DiT-based models comprise \textit{Flux.1-dev} \cite{flux1ai2024}, \textit{In-Context LoRA} (IC-LoRA) \cite{IC-Lora}, and \textit{Regional-Prompting FLUX} (RPF) \cite{chen2024training}. 

Notably, SKD, IC-LoRA, and RPF support \textbf{Joint Generation (JG)} by modeling inter-step context, enabling the generation of all step images in a single forward pass. In contrast, other models generate each step independently. As shown in \textbf{Tab.\ref{tab:t2i}}, we evaluate all models on \textbf{Goal Faithfulness (GF)}, \textbf{Step Faithfulness (SF)}, \textbf{Cross-Step Consistency (CSC)}, \textbf{Ingredient Accuracy (IA)}, and \textbf{Interaction Faithfulness (IF)}.

As shown in Fig.\ref{fig:vis}, among JG-capable models, IC-LoRA and RPF tend to produce overly smooth and temporally coherent sequences, but often suffer from \textbf{Position Misalignment}, where the visual content does not accurately correspond to the respective step descriptions.  Although SKD benefits from recipe-specific training data and exhibits relatively good consistency, it is limited to generating only six steps per recipe and still shows \textbf{Overly-Consistent} issues in some visualizations.

In contrast, non-JG models generate each step independently, which can cause \textbf{Ingredient Missing}—previously introduced but still relevant ingredients are omitted if not mentioned in the current step—and \textbf{Ingredient Inconsistency}, where the same ingredient appears differently across steps. As shown in Fig.~\ref{fig:vis}, long spaghetti noodles may become short pasta, breaking visual coherence and reducing procedural fidelity.

\begin{table}
  \centering
  \tabcolsep=6pt % 增加列间距
  \small         % 使用较小字体以适应宽度
    \caption{Experiments on Text-to-Image Models.}
  \begin{tabular}{@{}llccccc@{}}
    \toprule
    % \rowcolor{gray!20} % 表头浅灰色背景
    \textbf{Method} &\textbf{JG}&\textbf{GF}$\uparrow$& \textbf{SF}$\uparrow$& \textbf{CSC}$\downarrow$& \textbf{IA(\%)}$\uparrow$& \textbf{IF}$\uparrow$\\
    \midrule
    SD1.5~\cite{rombach2022high} &  \XSolidBrush &26.84& 28.40& 5.42&36.04 & 24.60\\
    SD2.1~\cite{stabilityai2022sd21} &  \XSolidBrush &26.88& 28.51& 7.54&39.86 & 24.95\\
    SDXL ~\cite{podell2023sdxl}&  \XSolidBrush &27.46& 29.37& 2.98&44.04 & 24.27\\
 SKD~\cite{menon2024generating} &  \Checkmark   &26.62& 28.53& 0.7&42.34 &24.71\\
    SD3.5 ~\cite{stabilityai2024sd35large}&  \XSolidBrush &27.42& 28.77& 2.97&45.27 & 24.11\\
    Flux.1-dev~\cite{flux1ai2024} &  \XSolidBrush &26.47& 28.31&3.47&38.88 & 24.06\\
    IC-LoRA~\cite{IC-Lora} &  \Checkmark   &26.07& 26.58& 9.03&24.28 & 24.17\\
    RPF~\cite{chen2024training} &  \Checkmark   &27.19& 25.99&8.73&31.13 & 24.59\\
    \bottomrule
  \end{tabular}

  \label{tab:t2i}
\end{table}
% \vspace{-2.5em}  % 你可以根据实际需要调节这个值

\textbf{Discussion on Recipe Image Generation.}
The task of recipe image generation aims to produce a sequence of visual illustrations based on a multi-step recipe text. These images should maintain visual continuity while accurately reflecting the mentioned ingredients and their interactions at each step. For models that support \textbf{Joint Generation}, where the entire recipe text is available in a single forward pass, there is inherent potential to ensure better semantic consistency across steps. However, the key challenge lies in \textbf{binding each step description to its correct spatial region} within the generated sequence and ensuring that each image corresponds faithfully to its respective textual instruction. In contrast, for \textbf{step-by-step generation models}, the primary difficulty is the lack of inter-step memory, which often leads to inconsistencies in background, ingredient presence, and visual context. To address this, such models may benefit from introducing a \textbf{memory mechanism} that retains information from previous steps—such as generated images, ingredient states, and contextual cues. However, this approach must carefully balance \textbf{computational overhead} in terms of time and memory usage. Moreover,  as shown in Tab.\ref{tab:t2i}, all existing models struggle to generate the correct interaction types between ingredients, making fine-grained depiction of ingredient interactions a critical open problem that future work must address.

\subsection{Recipe Video Generation}

We also evaluate the task of video generation, which is divided into two settings: (1) generating a video for each step based solely on the step-wise textual description, and (2) generating step-wise videos conditioned on both the textual description and the corresponding step image. We randomly select 384 videos for test. The specific result is in Tab.\ref{tab:video}. Visualization is in \href{https://wenbin08.github.io/RecipeGen/}{\textcolor{skyblue}{Project page}}.

% \begin{table}
%   \centering
%   \tabcolsep=6pt % 增加列间距
%   \small         % 使用较小字体以适应宽度
%   % \renewcommand{\arraystretch}{1.2} % 增加行间距
%     \caption{Experiments on Video Generation models.}
%   \begin{tabular}{@{}llccccc@{}}
%     \toprule
%     % \rowcolor{gray!20} % 表头浅灰色背景
%     \textbf{Method}  &\textbf{TASK}&\textbf{GF}$\uparrow$& \textbf{SF}$\uparrow$& \textbf{CSC}$\downarrow$& \textbf{IA (\%)}$\uparrow$& \textbf{IF}$\uparrow$\\
%     \midrule
%     Hunyuan \cite{kong2024hunyuanvideo} &T2V&29.66&29.75 & 0.02& 46.02& 29.69\\
%     Hunyuan \cite{kong2024hunyuanvideo} &I2V&28.56&29.24 & 0.15& 51.42& 28.73\\
%     Opensora \cite{opensora} &T2V&30.88&30.59 & 0.03& 56.20& 30.21\\
%  Opensora \cite{opensora}&I2V&29.58&29.14 & 0.16& 55.67&29.40\\
%   \bottomrule
%   \end{tabular}

%   \label{tab:video}
% \end{table}
% % \vspace{-2.5em}  % 你可以根据实际需要调节这个值

\textbf{Discussion on Recipe Video Generation. }In current video generation models, using key frames for the I2V task does not perform as well as pure text generation (T2V). This is because, in recipe videos, the step text describes instructional actions not the image, and key frame images do not strictly match those actions. This mismatch makes I2V less effective, as the model struggles to extract information from key frames that aligns with the current step’s instructions. Key frames are taken at a single moment and cannot fully capture the dynamic process of an action, which leads to errors in temporal coherence and step consistency in the generated video. In contrast, pure text generation relies only on abstract instructions, which more clearly guide the model to capture action meaning and the correct order of steps, resulting in better performance on GF, SF, and CSC metrics.

% \begin{table}
%   \centering
%   \tabcolsep=6pt
%   \small
%   \caption{Experiments on Video Generation models.}
%   \begin{tabular}{@{}llccccc@{}}
%     \toprule
%     \textbf{Method} & \textbf{TASK} & \textbf{GF}$\uparrow$ & \textbf{SF}$\uparrow$ & \textbf{CSC}$\downarrow$ & \textbf{IA (\%)}$\uparrow$ & \textbf{IF}$\uparrow$ \\
%     \midrule
%     \multirow{2}{*}{Hunyuan \cite{kong2024hunyuanvideo}} 
%     & T2V & 29.66 & 29.75 & 0.02 & 46.02 & 29.69 \\
%     & I2V & 28.56 & 29.24 & 0.15 & 51.42 & 28.73 \\
%     \midrule
%     \multirow{2}{*}{Opensora \cite{opensora}} 
%     & T2V & 30.88 & 30.59 & 0.03 & 56.20 & 30.21 \\
%     & I2V & 29.58 & 29.14 & 0.16 & 55.67 & 29.40 \\
%     \bottomrule
%   \end{tabular}
%   \label{tab:video}
% \end{table}

\begin{table}[t]
  \centering
  \small
  \setlength{\tabcolsep}{6pt}
  \renewcommand{\arraystretch}{1.2}
  \caption{Comparison of Video Generation Models. Hunyuan means HunyuanVideo model.}
  \begin{tabular}{llccccc}
    \toprule
    \textbf{Method} & \textbf{Task} & \textbf{GF}$\uparrow$ & \textbf{SF}$\uparrow$ & \textbf{CSC}$\downarrow$ & \textbf{IA (\%)}$\uparrow$ & \textbf{IF}$\uparrow$ \\
    \midrule
    \multirow{2}{*}{\texttt{Hunyuan}~\cite{kong2024hunyuanvideo}} 
    & T2V & 29.66 & 29.75 & 0.02 & 46.02 & 29.69 \\
    & I2V & 28.56 & 29.24 & 0.15 & 51.42 & 28.73 \\
    \midrule
    \multirow{2}{*}{\texttt{Open-Sora}~\cite{opensora}} 
    & T2V & 30.88 & 30.59 & 0.03 & 56.20 & 30.21 \\
    & I2V & 29.58 & 29.14 & 0.16 & 55.67 & 29.40 \\
    \bottomrule
  \end{tabular}
  \label{tab:video}
\end{table}

% \subsection{RecipeQA Task}
% We also evaluated various multimodal large models on recipe question answering and several fine-grained classification tasks; details can be found in \href{https://huggingface.co/datasets/RUOXUAN123/RecipeGen}{\textcolor{skyblue}{Project page}}.
% \subsection{Cooking }

\section{Conclusion}

In this work, we present RecipeGen, the first step-aligned, multimodal benchmark for real-world recipe generation. By providing richly annotated text, image, and video data, RecipeGen addresses the lack of comprehensive datasets and metrics in food computing. We believe that our benchmark can foster further advancements in food computing, particularly in real-world, interactive culinary applications.

\section{Usage and License}

All data used in the RecipeGen benchmark is collected from publicly available sources. RecipeGen is released under the \textbf{Creative Commons Attribution-NonCommercial 4.0 International (CC BY-NC 4.0)} license. This license permits sharing and adaptation of the dataset for \textbf{non-commercial} purposes, provided appropriate credit is given. \textbf{Commercial use is strictly prohibited} under the terms of this license. The dataset is publicly available under the CC BY-NC 4.0 license at \href{https://huggingface.co/datasets/RUOXUAN123/RecipeGen}{\textcolor{skyblue}{HuggingFace}}.

\clearpage
\clearpage
\bibliographystyle{ACM-Reference-Format}
\bibliography{sample-base}

\appendix
\clearpage
\clearpage
% \input{sec/supplementary}

% % \section{Research Methods}








\end{document}